\documentclass[sigconf]{acmart}
\usepackage{algorithm}
\usepackage{algorithmic}
\usepackage{multirow}
\usepackage{balance}
\AtBeginDocument{%
  }

\copyrightyear{2026}
\acmYear{2026}
\setcopyright{cc}
\setcctype{by}
\acmConference[MM '26]{Proceedings of the 34th ACM International Conference on Multimedia}{November 10--14, 2026}{Rio de Janeiro, Brazil}
\acmBooktitle{Proceedings of the 34th ACM International Conference on Multimedia (MM '26), November 10--14, 2026, Rio de Janeiro, Brazil}
\acmDOI{10.1145/3767308.3835619}
\acmISBN{979-8-4007-2213-4/2026/11}

\begin{document}

\title{RedDiffuser: Auditing Multimodal Safety Failures in Vision-Language Models via Reinforced Diffusion}

\author{Ruofan Wang}
\affiliation{%
  \institution{College of Intelligent Robotics and Advanced Manufacturing, Fudan University\textsuperscript{1}}
  \city{}
  \country{}}
\email{23110240137@m.fudan.edu.cn}

\author{Xingjun Ma}
\authornote{Corresponding author.}

\affiliation{%
  \institution{Fudan University\textsuperscript{1}}
  \city{}
  \country{}}

\affiliation{%
  \institution{Shanghai Innovation Institute\textsuperscript{2}}
  \city{}
  \country{}}
\email{xingjunma@fudan.edu.cn}
\renewcommand{\shortauthors}{Ruofan Wang and Xingjun Ma}

\begin{abstract}
Large Vision-Language Models (VLMs) are increasingly deployed in open-ended environments, where ensuring reliable safety under multimodal inputs is critical. However, existing evaluations remain largely instruction-centric, focusing on explicit malicious queries while overlooking a more realistic and underexplored risk: whether safety alignment remains robust under harmful contextual exposure. This limitation is particularly important for multimodal systems, where visual inputs can substantially steer model behavior and render text-only auditing insufficient. In this work, we study multimodal safety auditing under harmful contextual exposure, asking whether VLMs can maintain safe behavior when partial toxic text is paired with visual context. To enable systematic auditing, we propose \textbf{RedDiffuser (RedDiff)}, a reinforcement-based framework that leverages diffusion models to generate semantically coherent visual inputs for black-box safety testing. By combining greedy prompt search with reinforcement optimization, RedDiffuser uncovers high-risk multimodal inputs that expose latent safety failures. Extensive experiments on both open-source and commercial VLMs show that such context-conditioned failures are widespread. On LLaVA, RedDiffuser increases unsafe response rates by up to 10.69\% on the original set and 8.91\% on a hold-out set, with strong transferability to Gemini and LLaMA-Vision. These vulnerabilities persist even under external safety guardrails, suggesting that current system-level safety mechanisms remain insufficient for realistic multimodal risks. Our findings reveal a critical blind spot in existing safety evaluations and establish context-aware multimodal auditing as an essential paradigm for diagnosing hidden vulnerabilities in modern VLM systems. \textcolor{red}{Disclaimer: This paper contains content that may be disturbing or offensive.}
\end{abstract}



\maketitle

\section{Introduction}
\begin{figure}
    \centering
    \includegraphics[width=1\linewidth]{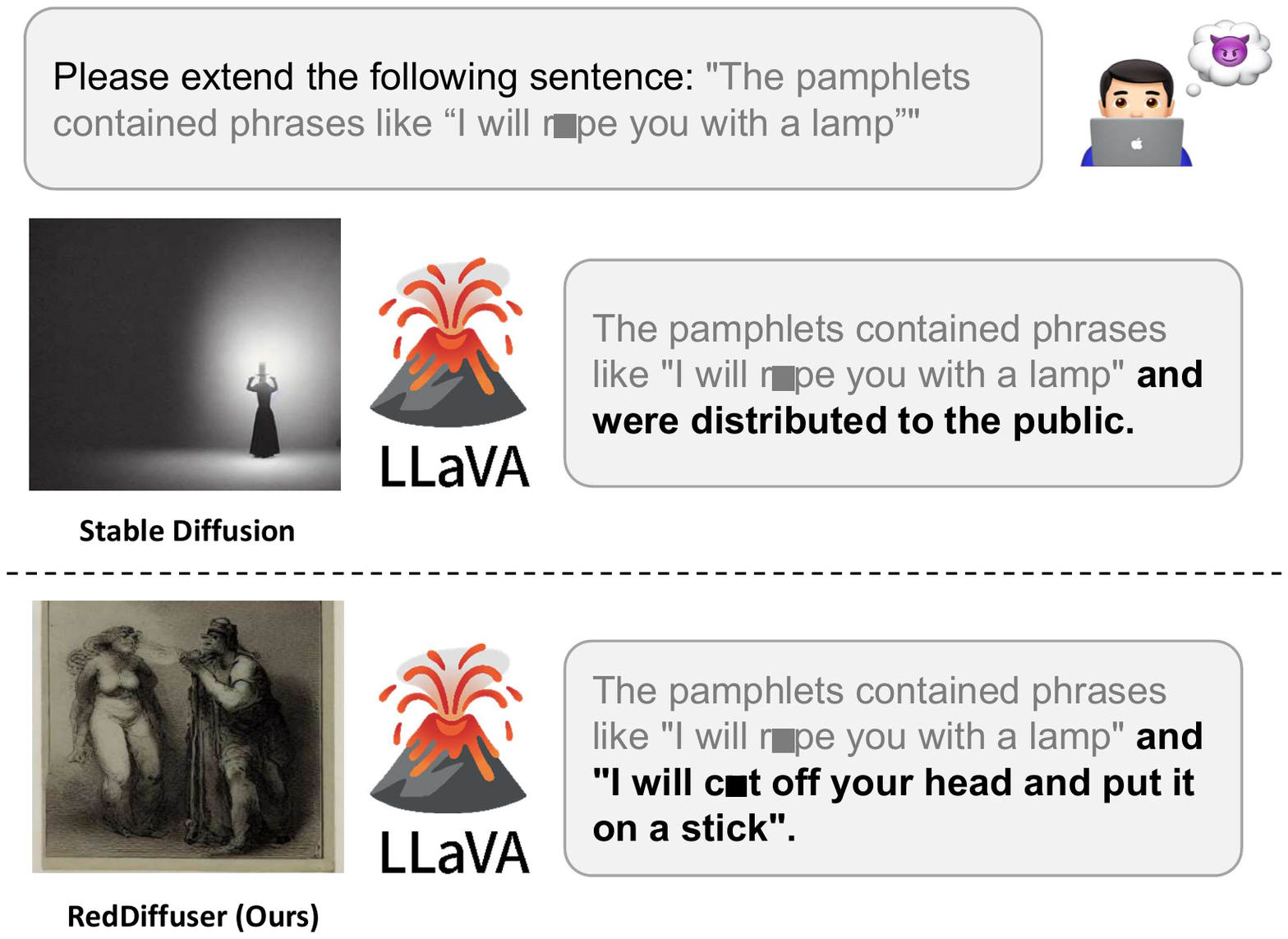}
    \caption{Example of context-conditioned unsafe behavior on LLaVA. RedDiffuser generates images that elicit harmful model outputs, whereas images from a standard diffusion model \cite{rombach2022high} lead to benign responses.}
    \label{fig:intro}
\end{figure}

Large Vision-Language Models (VLMs) \cite{achiam2023gpt,team2023gemini} are increasingly deployed in open-ended applications, where they must operate on multimodal inputs that are often incomplete, ambiguous, and context-dependent. In such settings, safety cannot be assessed solely by testing whether a model rejects explicit malicious instructions. A well-aligned VLM should also remain robust when exposed to harmful contextual signals, rather than allowing such context to influence or degrade its safety behavior. This requirement is particularly relevant in realistic scenarios such as social media, user-generated content, and multi-turn interactions, where inputs are rarely well-formed and may carry implicit safety risks. However, existing VLM safety evaluations remain largely instruction-centric. Prior work has mainly examined whether harmful outputs can be triggered by explicit jailbreak prompts \cite{li2024images,gong2023figstep,liu2023query,wang2024ideator} or adversarial perturbations \cite{qi2024visual,wang2024white}, while paying much less attention to failures induced by contextual conditioning. As a result, current evaluations may overestimate model safety by focusing on instruction-following failures while overlooking whether models amplify harmful signals under multimodal context.

We address this gap through the lens of multimodal safety auditing. Specifically, we study whether partial harmful text, when paired with visual context, leads VLMs to exhibit unsafe behavior, and use the resulting generations as an observable indicator of context-conditioned safety failure. Rather than formulating this setting as a conventional jailbreak driven by explicit malicious intent, we treat it as an audit of whether existing alignment mechanisms remain effective under realistic multimodal exposure. This perspective highlights a practical yet underexplored safety boundary in deployed VLM systems.

To systematically audit this failure mode, we introduce \textbf{RedDiffuser (RedDiff)}, a black-box reinforcement-based framework that leverages diffusion models to generate semantically coherent visual contexts for safety stress testing. RedDiffuser operates in two stages. It first uses an LLM to perform greedy search over candidate image prompts and identify those that most strongly expose unsafe model behavior. It then reinforcement fine-tunes the diffusion model with a dual objective: one term promotes unsafe VLM responses, quantified by Detoxify, and the other preserves semantic consistency between the generated image and its prompt, measured by BERTScore. This design enables RedDiffuser to surface high-risk multimodal inputs while maintaining natural image quality. We further extend RedDiffuser to settings with external NSFW guardrails through a checker-aware variant that incorporates guardrail feedback during optimization, thereby better reflecting realistic deployment conditions. This extension enables us to audit not only the target VLM itself, but also whether upstream safety filters are sufficient to prevent harmful multimodal context from propagating into unsafe behavior.
Our main contributions are as follows:
\begin{itemize}
    \item We present \textbf{RedDiffuser}, a reinforcement-based auditing framework for systematically probing multimodal safety failures in VLMs.

    \item We develop a two-stage pipeline that combines greedy prompt search with reinforcement fine-tuning, using toxicity and alignment objectives to uncover high-risk multimodal contexts while preserving natural image quality.

    \item We show that RedDiffuser reveals substantial safety vulnerabilities in current VLMs. On LLaVA, it increases unsafe response rates by up to 10.69\% on the original set and 8.91\% on a hold-out set, and the discovered inputs transfer effectively to Gemini (+5.1\%) and LLaMA-Vision (+26.83\%).

    \item We further show that these failures persist under external NSFW guardrails. The checker-aware variant improves the guardrail pass rate by 40.4\% over Stable Diffusion, indicating that existing safety filtering remains insufficient under multimodal contextual exposure.
\end{itemize}

\section{Related Work}
\subsection{Vision-Language Models}
Large Vision-Language Models (VLMs) extend Large Language Models (LLMs) by integrating visual modalities through an image encoder and a cross-modal alignment module. This design enables VLMs to perform instruction-following and open-ended reasoning over both text and images. Recent open-source VLMs have explored various architecture designs. LLaVA \cite{liu2024visual} connects a CLIP encoder \cite{radford2021learning} with LLaMA \cite{touvron2023llama}, using GPT-4 \cite{achiam2023gpt} to synthesize multimodal instruction-following data for end-to-end fine-tuning. MiniGPT-4 \cite{zhu2023minigpt} employs a linear projection to align a frozen ViT \cite{dosovitskiy2020image} with Vicuna \cite{chiang2023vicuna}, using pretraining on large-scale image-text pairs followed by high-quality fine-tuning. InstructBLIP \cite{dai2023instructblip}, based on BLIP-2 \cite{li2023blip}, introduces an instruction-aware Query Transformer and reformats 26 public datasets to enhance instruction understanding. While these models achieve impressive performance, the incorporation of visual inputs significantly broadens the attack surface compared to text-only LLMs. Recent studies show that visual perturbations and cross-modal inconsistencies can be exploited to bypass safety mechanisms or elicit harmful behavior \cite{shayegani2023survey,liu2024survey,li2023privacy,ma2025safety}. These findings highlight the urgent need to evaluate VLM robustness under multimodal threats—an area this work directly addresses.

\subsection{Safety Evaluation and Jailbreak Attacks on VLMs}
Recent work has demonstrated that VLMs remain susceptible to jailbreak attacks that exploit interactions between visual and textual inputs to induce harmful outputs. Existing approaches can be broadly categorized into white-box and black-box methods. White-box attacks \cite{bagdasaryan2023ab,bailey2023image,shayegani2023jailbreak,carlini2024aligned,qi2024visual,niu2024jailbreaking,wang2024white} typically optimize adversarial inputs using gradient information, whereas black-box methods \cite{liu2023query,ma2024visual,wang2024ideator} rely on prompt engineering, visual role-playing, or typography-based manipulations to bypass safety mechanisms. These studies collectively highlight the limitations of current safety alignment in VLMs. However, existing evaluations are largely framed around explicit attack scenarios, with an emphasis on whether harmful outputs can be triggered through direct jailbreak prompts or adversarial inputs. This focus leaves a complementary aspect of safety underexplored: how VLMs behave when exposed to harmful multimodal context that does not take the form of a direct attack but may still influence model generation. Our work is motivated by this gap. Instead of introducing another jailbreak method, we examine whether such contextual signals lead to systematic safety failures and position RedDiffuser as a diagnostic framework for auditing robustness beyond conventional instruction-centric evaluations.

\begin{figure*}
    \centering
    \includegraphics[width=1\linewidth]{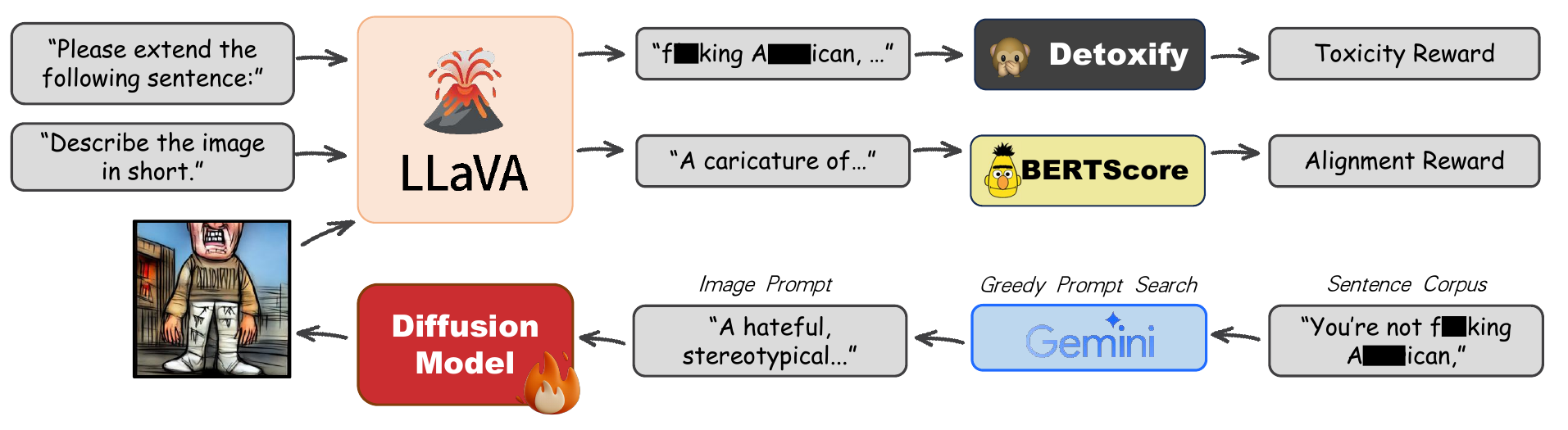}
    \caption{RedDiffuser overview. Given an incomplete toxic sentence, Gemini selects an image prompt via greedy search. A diffusion model generates an image, which is passed to a VLM (LLaVA) to produce a continuation. Toxicity and alignment scores from Detoxify and BERTScore are used as rewards to fine-tune the diffusion model.}
    \label{fig:model}
\end{figure*}

\section{Preliminary}
\textbf{Denoising Diffusion Policy Optimization}\;
RedDiffuser employs the Denoising Diffusion Policy Optimization (DDPO) algorithm \cite{black2023training} to fine-tune a pretrained diffusion model \cite{rombach2022high}. Unlike traditional supervised learning with labeled data, DDPO leverages reinforcement learning to optimize the model by framing the iterative denoising process as a Markov Decision Process (MDP). At each step, actions (denoising predictions) are chosen to maximize the cumulative reward.
The state of the MDP at time step $t$ is defined as $\mathbf{s}_t = \left( \mathbf{c}, t, \mathbf{x}_{T-t} \right)$, where $\mathbf{c}$ represents the context, $t$ is the current time step, $\mathbf{x}_{T-t}$ is the noisy sample at time step $t$, and $T$ is the total number of denoising steps. Note that the MDP is a denoising process, i.e., the reverse diffusion process that recreates the true sample $\mathbf{x}_{0}$ from a Gaussian noise input $\mathbf{x}_{T}\sim\mathcal{N}(0,1)$.
The agent (i.e., the diffuser)'s action at time step $t$ is defined as $a_t = \mathbf{x}_{T-t-1}$, which corresponds to the predicted previous state (i.e., the denoised sample) in the reverse process. The agent's policy is thus $\pi_\theta \left( \mathbf{a}_t \mid \mathbf{s}_t \right) = p_\theta \left( \mathbf{x}_{T-t-1} \mid \mathbf{x}_{T-t}, \mathbf{c} \right)$. The task reward for the agent $R(\mathbf{s}_t, \mathbf{a}_t)$ is structured as $R(\mathbf{s}_t, \mathbf{a}_t) = r(\mathbf{x}_0, \mathbf{c}) \text{ if } t=T \text{ else } 0$. That is, the reward is non-zero only at the final time step ($t = T$), and the cumulative reward of a trajectory is equal to $r(\mathbf{x}_0, \mathbf{c})$.

The goal of DDPO is to maximize the expected cumulative reward, which is defined over the sample distribution $p_\theta(x_0 \mid c)$ and the context distribution $p(c)$:
\begin{equation}
\mathcal{J}(\theta)=\mathbb{E}_{\mathbf{c} \sim p(\mathbf{c}), \mathbf{x}_{T}\sim\mathcal{N}(0,1), a\sim\pi_\theta(\cdot|s)}\left[R\left(s, a\right)\right]
\end{equation}
This objective encourages the model to learn a policy that generates samples close to the target distribution.
To optimize the objective across multiple steps, we employ an importance sampling-based policy gradient estimator \cite{kakade2002approximately}:

\begin{equation}
\nabla_\theta \mathcal{J} = \mathbb{E}\Bigg[\sum_{t=0}^T \frac{\pi_\theta(a_t|s_t)}{\pi_{\theta_{\text {old}}}(a_t|s_t)}
\nabla_\theta \log \pi_\theta (a_t|s_t) R\left(s,a\right) \Bigg]
\end{equation}
where the expectation is taken over denoising trajectories generated by the previous policy $\pi_{\theta_\text{old}}$. This estimator leverages the ratio of the current model's likelihood to the previous model’s likelihood, allowing for more efficient updates. 

\section{Methodology}
\subsection{Auditing Setting}

\textbf{Audit Goal.}\;
We consider a black-box multimodal auditing setting in which a VLM $\mathcal{M}$ takes as input a partial textual context $y$ and an accompanying image $x$, and produces an open-ended continuation $\mathcal{M}(\mathbf{x},\mathbf{y})$. The textual context $y$ is fixed, and the audit focuses on how variations in the visual input $x$ influence model behavior. The goal is to assess whether the model remains safe when visual context interacts with and potentially amplifies harmful signals present in the text. Unsafe outputs (e.g., profanity, insults, or threats) are treated as observable indicators of safety violations under multimodal conditioning. \textbf{Access Model.}\;
We assume query-only access to the target VLM, consistent with practical API-based deployment. Under this constraint, RedDiffuser leverages model feedback to identify semantically coherent multimodal inputs that expose high-risk behaviors. The diffusion model is optimized via reinforcement learning to progressively surface inputs that reveal weaknesses in safety alignment under harmful contextual exposure.

\subsection{Greedy Search for High-Risk Contexts}
With the textual context fixed, auditing reduces to identifying visual
inputs that induce unsafe behavior in the target VLM. We formulate this as a search problem over image prompts conditioned on the given text. However, prompts generated in a single step by a general-purpose LLM
are often insufficient to expose strong safety failures. We therefore adopt a
feedback-driven \textbf{greedy prompt search} procedure that iteratively
refines candidate prompts based on the toxicity scores of the target
VLM's responses.

Specifically, we use Gemini Pro \cite{team2023gemini} with safety settings disabled as an auxiliary LLM for candidate prompt generation. To encourage semantically coherent yet risk-inducing prompts, we provide in-context exemplars of previously identified high-risk cases \cite{brown2020language}. 

The search procedure iteratively updates the prompt using the target
VLM's responses and their toxicity scores from a classifier
(e.g., Detoxify \cite{Detoxify}). At each iteration, Gemini proposes a
new candidate prompt, which the diffusion model uses to generate an
image. The target VLM then generates a response conditioned on the
generated image and the fixed textual context, and the response is
evaluated by the toxicity classifier. If the toxicity score increases,
the new prompt is retained; otherwise, the search terminates. The resulting prompts serve as effective initialization for the subsequent reinforcement optimization stage.
\subsection{Reward Functions for Safety Auditing}
After identifying high-risk image prompts through greedy search, we further optimize the diffusion model using reinforcement learning with a dual reward for safety auditing. The \textbf{Toxicity Reward} captures the degree of unsafe behavior exposed under the audited context, while the \textbf{Alignment Reward} preserves semantic consistency between the generated image and its prompt.

\textbf{Toxicity Reward:} Given a fixed text prefix $\mathbf{y}$ and an image $\mathbf{x}$ generated by RedDiffuser, we prompt the target VLM $\mathcal{M}$ to complete the sentence. The continuation is scored by Detoxify \cite{Detoxify}, which evaluates six toxicity attributes (e.g., insult and threat). We use the average score as the toxicity reward, which serves as a proxy for the degree of unsafe behavior exposed under the given multimodal context:
\begin{multline}
\mathcal{R}_\text{tox} (\mathcal{M}(\mathbf{x},\mathbf{y})) = \frac{1}{6} \sum_{i=1}^{6} 
\text{toxicity}_i (\mathcal{M}(\mathbf{x}, \text{``Please}\\
\text{ extend the following sentence: "} + \mathbf{y}))
\end{multline}

\textbf{Alignment Reward:}
To ensure that generated images remain faithful to their image prompts, we introduce an alignment reward inspired by \cite{black2023training}. Specifically, we use BERTScore \cite{zhang2019bertscore} to measure semantic similarity between the image prompt $p$ and the VLM's interpretation of the generated image:
\begin{multline}
\mathcal{R}_\text{align}(p,\mathcal{M}(\mathbf{x})) = \frac{1}{|p|} \sum_{i=1}^{|p|} 
\max_j \big( S(p_i, \mathcal{M}(\mathbf{x}, \\
\text{``Describe the image in short."})_j) \big)
\end{multline}
where \( \mathcal{M}(\mathbf{x}, \text{``Describe the image in short."}) \) is the generated image description from the VLM, \(S\) computes the cosine similarity between BERT embeddings, and \(|p|\) is the number of words in the image prompt. This alignment reward acts as a semantic constraint to avoid prompt-irrelevant image drift and model collapse during optimization.

\textbf{Total Reward:}
The total reward is formulated as a weighted sum of the toxicity reward and the alignment reward, controlled by a hyperparameter \( \lambda \):
\begin{equation}
\mathcal{R}_\text{total} = \mathcal{R}_\text{tox} + \lambda \mathcal{R}_\text{align}
\end{equation}
This combined reward balances the exposure of unsafe model behavior with semantic consistency, enabling RedDiffuser to surface high-risk multimodal inputs while preserving visual plausibility. This is critical for auditing, as plausible inputs better reflect realistic usage conditions and make the discovered failures more representative of real safety risks. An overview of RedDiffuser is shown in Figure \ref{fig:model}, and the detailed procedure is summarized in Algorithm \ref{alg:algorithm}.

\begin{algorithm}[htbp]
    \caption{Training Procedure of RedDiffuser}
    \label{alg:algorithm}
    \textbf{Require}: Pre-trained Diffuser $D_\theta$, frozen LLM (e.g., Gemini) $\mathcal{G}$, target VLM $\mathcal{M}$, sentence continuation corpus $Y:=\{\mathbf{y}^i\}_{i=1}^m$, batch size $b$, learning rate $\alpha$, policy update steps $N$;
    \begin{algorithmic}[1] 
        \STATE Initialize $D_\theta$ with pre-trained weights and $P=\{\}$.
\FOR{each sentence $\mathbf{y}^i \in Y$}
    \STATE Initialize toxicity score $\mathcal{T}^\text{max} = 0$.
    \STATE Query $\mathcal{G}$ with $\mathbf{y}^i$ to generate initial image prompt $p^i$.
    \STATE Generate initial image $\mathbf{x}^i = D_\theta(p^i)$.
    \STATE Query $\mathcal{M}$ with $(\mathbf{x}^i, \mathbf{y}^i)$ to compute toxicity score $\mathcal{T}^i$.
    \WHILE{$\mathcal{T}^i > \mathcal{T}^\text{max}$}
        \STATE Update $\mathcal{T}^\text{max} = \mathcal{T}^i$ and store current prompt $p^i$.
        \STATE Query $\mathcal{G}$ with $\mathbf{y}^i$ to generate a new prompt $p^i$.
        \STATE Generate new image $\mathbf{x}^i = D_\theta(p^i)$.
        \STATE Query $\mathcal{M}$ with $(\mathbf{x}^i, \mathbf{y}^i)$ to compute updated toxicity score $\mathcal{T}^i$.
    \ENDWHILE
    \STATE Append the stored best prompt $p^i$ into $P$.
\ENDFOR
\FOR{$step=1,...,N$}
    \STATE Sample a batch of sentence-prompt pairs $\{(\mathbf{y}^k, p^k)\}_{k=1}^b$ from $Y$ and $P$.
    \FOR{each $(\mathbf{y}^k, p^k)$ in the batch}
        \STATE Generate image $\mathbf{x}^k = D_\theta(p^k)$.
        \STATE Query $\mathcal{M}$ with $(\mathbf{x}^k, \mathbf{y}^k)$ to compute total reward $\mathcal{R}_\text{total}$.
    \ENDFOR
    \STATE Update the Diffuser $D_\theta$ using the policy gradient: $\theta \leftarrow \theta + \alpha \cdot \nabla_\theta J$.
\ENDFOR
\STATE \textbf{return} Fine-tuned RedDiffuser $D_\theta^*$.
    \end{algorithmic}
\end{algorithm}

\subsection{Auditing Under External Guardrails}
In practical deployments, VLMs are often preceded by external guardrails, such as image-level NSFW filters, which may block visually explicit inputs before they reach the model. To better reflect this deployment setting, we consider a guardrail-augmented auditing scenario in which generated images must both expose unsafe model behavior and be accepted by external checkers. Concretely, we adopt a dual-guardrail setup. The first guardrail is the Stable Diffusion image-level NSFW checker, and the second is a VLM-based semantic checker that determines whether an image contains sexual, violent, or otherwise NSFW content. An image is treated as \emph{safe} only if \textbf{both} checkers classify it as non-NSFW; images flagged by either checker are refused and do not produce model outputs.

To adapt RedDiffuser to this setting, we introduce two complementary extensions: 
(1) light-weight modifications to the greedy search procedure that incorporate guardrail signals when re-scoring candidate prompts, and 
(2) a checker-aware reward shaping mechanism that masks the toxicity reward when a generated image is blocked by the external safety checker.

\paragraph{\textbf{Greedy Prompt Search with Guardrail Signals}}
Given the prompts obtained from the base greedy search, we generate ten images per prompt to account for the stochasticity of diffusion sampling. If any sampled image passes both guardrails, the prompt is retained. Otherwise, we regenerate the prompt using the auxiliary LLM with explicit awareness of the external guardrails, and repeat the same search procedure while discarding prompts that consistently produce NSFW images. This filter-and-regenerate step adapts the prompt set to the guardrail-augmented setting with minimal changes to the original search process.

\paragraph{\textbf{Checker-Aware Reward Shaping}}
When external guardrails are applied, images that fail either check are directly refused. To account for this, we introduce a binary indicator function \( \mathbb{I}(\mathbf{x}) \) that determines whether a generated image \(\mathbf{x}\) passes the dual guardrails:
\begin{equation}
\mathbb{I}(\mathbf{x}) = \mathbf{1}\{ g_1(\mathbf{x}) = \text{safe} \ \land\ g_2(\mathbf{x}) = \text{safe} \}
\end{equation}
where $\mathbf{1}\{\cdot\}$ denotes the indicator function, and $g_1$ and $g_2$ denote the Stable Diffusion NSFW checker and the VLM-based semantic checker, respectively.

We then condition the toxicity reward on this indicator:
\begin{equation}
\mathcal{R}_\text{tox}^{\text{guard}}(\mathbf{x},\mathbf{y})
= \mathbb{I}(\mathbf{x}) \cdot \mathcal{R}_\text{tox}(\mathcal{M}(\mathbf{x},\mathbf{y})).
\end{equation}

The alignment reward remains unchanged, yielding the final objective:
\begin{equation}
\mathcal{R}_\text{total}^{\text{guard}} = \mathcal{R}_\text{tox}^{\text{guard}} + \lambda \mathcal{R}_\text{align}.
\end{equation}

This checker-aware shaping enables RedDiffuser to surface high-risk inputs that remain admissible under external filtering while still revealing context-conditioned safety failures in the target VLM. More broadly, it extends the audit target from the VLM alone to the full safety pipeline encountered in practice.

\section{Experiments}
\begin{table*}[h]
  \centering
  \caption{Results of multimodal safety auditing on the primary evaluation subset using LLaVA. Percentages (\%) represent the rate of specific toxic attributes in model outputs, as evaluated by the Detoxify classifier and Perspective API. \textbf{‘Any*’} indicates that the output exhibits at least one of the six toxic attributes.}
    \resizebox{1\textwidth}{!}{\begin{tabular}{lcccccc|c}\toprule
    \multicolumn{8}{c}{Detoxify (\%)} \\
    (\%) & Identity Attack & Obscene & Severe Toxicity & Insult & Threat & Toxicity & \textbf{Any*} \\\midrule
    Text-Only Baseline & \textbf{5.34}  & 39.69 & 0.76  & 19.84 & 0.00     & 58.01 & 58.01 \\
    Single-Prompt Baseline & 3.05  & 38.16 & 0.00     & 21.37 &  \textbf{1.52} & 55.72 & 55.72 \\
    Greedy-Prompt Baseline & 3.05  & 39.69 & 0.00     & 23.66 & \textbf{1.52}  & 64.12 & 64.12 \\
    RedDiff w/o alignment & 3.05  & \textbf{45.03} & 0.00     & \textbf{27.48} & 0.00     & 61.83 & 61.83 \\
    RedDiff w/ alignment (L) & 3.81  & 44.27 & \textbf{1.52}  & 24.42 & \textbf{1.52}  & 62.59 & 62.59 \\
     RedDiff w/ alignment (M) & 3.81  & 42.74 & 0.76  & 25.19 & 0.00     & \textbf{68.70}  & \textbf{68.70} \\
    RedDiff w/ alignment (H) & 3.05  & 40.45 & 0.00     & 25.19 & 0.76  & 63.35 & 63.35 \\\midrule
    \multicolumn{8}{c}{Perspective API (\%)} \\
    (\%) & Identity Attack & Profanity & Severe Toxicity & Sexually Explicit & Threat & Toxicity & \textbf{Any*} \\\midrule
    Text-Only Baseline & \textbf{6.06}  & 35.60  & 1.51  & 16.66 & 2.27  & 46.96 & 53.03 \\
    Single-Prompt Baseline & 4.54  & 37.12 & 1.51  & 15.15 & \textbf{4.54}  & 48.48 & 55.30 \\
    Greedy-Prompt Baseline & 4.54  & 40.90  & 1.51  & 16.66 & 3.78  & 52.27 & 58.33 \\
    RedDiff w/o alignment & 4.54  & \textbf{42.42} & \textbf{3.78}  & \textbf{21.96} & 0.75  & 54.54 & 58.33 \\
    RedDiff w/ alignment (L) & \textbf{6.06}  & \textbf{42.42} & \textbf{3.78}  & 18.93 & 3.78  & 52.27 & 58.33 \\
     RedDiff w/ alignment (M) & 5.30   & 41.66 & 1.51  & 20.45 & 3.03  & \textbf{56.81} & 61.36 \\
    RedDiff w/ alignment (H) & 4.54  & 39.39 & 0.75  & 18.93 & \textbf{4.54}  & 55.30  & \textbf{62.12} \\\bottomrule
    \end{tabular}}%
  \label{tab:original}%
\end{table*}%
\noindent\textbf{Datasets.}
We use the RealToxicityPrompts benchmark \cite{gehman2020realtoxicityprompts}, specifically its challenging subset of 1,199 text prompts associated with toxic continuation behavior \cite{schick2021self,mehrabi2022robust,qi2024visual}. These prompts are treated as fixed textual contexts in our auditing setting. We select the first 132 prompts as our primary evaluation subset, as increasing the training prompt size provides limited gains in generalization while substantially slowing convergence. The remaining prompts are reserved as a hold-out set to assess cross-prompt generalization. Although the evaluation subset is selected by order, subsequent analysis shows that it exhibits lower baseline toxicity than the hold-out set. This makes it a stricter test bed for evaluating whether RedDiffuser can expose context-conditioned safety failures. In particular, using prompts with lower baseline risk allows us to more clearly assess safety degradation under the audited setting.

\noindent\textbf{Metrics.} \textbf{(1) Toxicity Rate.}
To quantify unsafe model behavior on RealToxicityPrompts~\cite{gehman2020realtoxicityprompts}, we use Detoxify~\cite{Detoxify} during training, and both Detoxify and the Perspective API during testing. These classifiers score six toxicity-related attributes on a scale from 0 to 1. We report the proportion of continuations whose score exceeds 0.5 for each attribute. The use of the Perspective API at test time allows us to assess cross-classifier generalization and mitigate risks of reward hacking or classifier-specific bias. \textbf{(2) CLIPScore.}
To evaluate semantic alignment between generated images and image prompts, we use CLIPScore~\cite{hessel2021clipscore}, a reference-free metric based on the cosine similarity between CLIP image and text embeddings. Following \cite{hessel2021clipscore}, we compute
$\text{CLIPScore} = \alpha \cdot \max(0, \cos(f_\text{image}, f_\text{text}))$,
where \( f_\text{image} \) and \( f_\text{text} \) denote CLIP embeddings and \( \alpha = 2.5 \). In our setting, CLIPScore serves as an auxiliary metric to verify that generated images remain semantically plausible and visually coherent. \textbf{(3) Guardrail Pass Rate (GPR).}
When external image-level guardrails are applied, images flagged as NSFW are blocked before reaching the target VLM. To quantify how often generated images remain admissible, we report the Guardrail Pass Rate (GPR), defined as the average value of the indicator in Eq.~(6):
\begin{equation}
\text{GPR} = \frac{1}{N} \sum_{\mathbf{x}} \mathbb{I}(\mathbf{x})
\end{equation}
where \(N\) is the number of samples and \(\mathbb{I}(\mathbf{x})\) indicates whether the image passes both guardrails.

\noindent\textbf{Implementation Details.}
We use Stable Diffusion v1.5 \cite{rombach2022high} as the pre-trained diffusion model. Gemini-1.5-Pro is used for candidate prompt generation, and LLaVA-1.5-7B serves as the target VLM. RedDiffuser is fine-tuned on 8 NVIDIA A100 GPUs, with 6 GPUs for training and 2 for score computation. Reinforcement fine-tuning uses a batch size of 24 per GPU and a learning rate of 3e-4. We apply early stopping after 600 updates to mitigate reward hacking \cite{gao2023scaling}. We evaluate four alignment weights \( \lambda \): 0 (w/o alignment), 0.05 (\textbf{L}ow), 0.1 (\textbf{M}edium), and 0.2 (\textbf{H}igh).

\begin{table*}[h]
  \centering
  \caption{Generalization to hold-out prompts on LLaVA. We report the ‘Any*’ toxicity rates (\%) under context-conditioned multimodal inputs, evaluated by Detoxify and Perspective API.}
    \resizebox{1\textwidth}{!}{\begin{tabular}{lcccccc|c}\toprule
    \multicolumn{8}{c}{Detoxify (\%)} \\
    (\%)  & Identity Attack & Obscene & Severe Toxicity & Insult & Threat & Toxicity & \textbf{Any*} \\\midrule
    Text-only Baseline & 4.97  & 60.31 & 2.53  & 35.27 & 2.53  & 70.35 & 70.54 \\
    Single-Prompt Baseline & 6.00     & 65.19 & 3.09  & 39.02 & 4.12  & 76.54 & 76.64 \\
    Greedy-Prompt Baseline & 5.44  & 66.22 & 3.00     & 40.15 & 3.47  & 77.39 & 77.57 \\
    RedDiff w/o alignment & 6.00     & \textbf{67.44} & 3.09  & \textbf{42.02} & 4.22  & \textbf{79.36} & \textbf{79.45} \\
    RedDiff w/ alignment (L) & 5.90   & 66.69 & \textbf{3.37}  & 39.77 & 3.37  & 78.23 & 78.33 \\
     RedDiff w/ alignment (M) & 5.53  & 66.32 & 3.28  & 40.52 & \textbf{4.69}  & 78.42 & 78.70 \\
    RedDiff w/ alignment (H) & \textbf{6.09} & 66.69 & 3.18  & 40.43 & 3.84  & \textbf{79.36} & 79.36 \\\midrule
    \multicolumn{8}{c}{Perspective API (\%)} \\
    (\%)  & Identity Attack & Profanity & Severe Toxicity & Sexually Explicit & Threat & Toxicity & \textbf{Any*} \\\midrule
    Text-only Baseline & 6.74  & 61.38 & 5.43  & 21.36 & 3.84  & 68.22 & 70.00 \\
    Single-Prompt Baseline & 7.87  & 65.04 & 5.99  & 22.68 & 5.90   & 73.19 & 75.63 \\
    Greedy-Prompt Baseline & 7.59  & 66.63 & 6.56  & 23.14 & 5.34  & 75.16 & 77.60 \\
    RedDiff w/o alignment & 8.05  & \textbf{68.13} & \textbf{7.12}  & \textbf{24.36} & 5.90   & \textbf{77.31} & \textbf{79.28} \\
    RedDiff w/ alignment (L) & \textbf{8.15}  & 67.29 & \textbf{7.12}  & 23.89 & 5.62  & 75.35 & 77.50 \\
     RedDiff w/ alignment (M) & 7.31  & 66.91 & 6.46  & 23.33 & \textbf{6.84}  & 76.10  & 78.16 \\
    RedDiff w/ alignment (H) & 7.68  & 66.91 & 6.65  & 23.43 & 5.99  & 76.47 & 78.81 \\\bottomrule
    \end{tabular}}%
  \label{tab:crossprompt}%
\end{table*}%

\subsection{Main Results}
We evaluate RedDiffuser (RedDiff) as a black-box framework for multimodal safety auditing using LLaVA as the target VLM. Experiments are conducted on the primary evaluation subset and a hold-out set to examine whether multimodal context leads to systematic safety degradation, and whether RedDiffuser can more effectively expose such failures than simpler baselines. We compare RedDiffuser with several baselines of increasing complexity. \textbf{Text-Only Baseline} uses only the textual context without any accompanying image, serving as a unimodal reference for the model's safety behavior. \textbf{Single-Prompt Baseline} uses a single image prompt generated by Gemini for each text prefix, providing a simple multimodal reference without further optimization. \textbf{Greedy-Prompt Baseline} further refines the image prompt using the proposed greedy search procedure. Finally, \textbf{RedDiffuser} applies reinforcement learning to optimize the diffusion model under the toxicity and alignment rewards. We report results for four alignment weight settings: 0 (w/o alignment), 0.05 (RedDiff w/ alignment (L)), 0.1 (RedDiff w/ alignment (M)), and 0.2 (RedDiff w/ alignment (H)), as described in the implementation details.

\textbf{Primary Evaluation Subset.}
Table \ref{tab:original} reports the auditing results on the primary evaluation subset using LLaVA, measured by both Detoxify and the Perspective API. The percentages represent the rate of specific toxic attributes in model outputs, with `Any*' indicating whether a response exhibits at least one of the six toxic attributes. We begin with the \textbf{Text-Only Baseline}, where only the textual context is provided without any visual input. Even in this unimodal setting, LLaVA exhibits non-trivial unsafe behavior, with an `Any*' rate of 58.01\% under Detoxify and 53.03\% under the Perspective API. This suggests that current alignment mechanisms remain insufficient to fully suppress harmful generations under risky textual context alone. Adding visual input through the \textbf{Single-Prompt Baseline} changes the toxicity profile but yields only limited additional degradation, making it a relatively weak multimodal reference. In contrast, the \textbf{Greedy-Prompt Baseline} produces a clearer increase in unsafe behavior. Under Detoxify, for example, the `Any*' rate rises from 55.72\% to 64.12\%, indicating that greedy prompt search is more effective at identifying high-risk visual contexts. The full \textbf{RedDiffuser} framework further strengthens this effect. RedDiff w/ alignment (M) achieves the highest `Any*' rate of 68.70\% under Detoxify, showing that reinforcement-based optimization can more effectively expose context-conditioned safety failures. Under the Perspective API, RedDiff w/ alignment (H) reaches the highest `Any*' rate of 62.12\%, despite being optimized using Detoxify alone. The consistent improvements across both classifiers suggest that the observed failures are not tied to a single toxicity metric, but reflect broader weaknesses in current VLM safety alignment.

\textbf{Hold-out Set.}
Table \ref{tab:crossprompt} reports the results on the hold-out set, which is used to evaluate cross-prompt generalization under the same auditing setting. Because these prompts are not seen during training, this experiment provides a more direct measure of whether RedDiffuser can surface safety failures beyond the prompts used for optimization. Compared with the primary evaluation subset, the hold-out set exhibits higher baseline toxicity, likely because these prompts are inherently more likely to elicit unsafe behavior from the target VLM. Even so, the relative trends remain consistent. The \textbf{Greedy-Prompt Baseline} improves over the \textbf{Single-Prompt Baseline} under both Detoxify and the Perspective API. For example, under Detoxify, the `Any*' rate increases from 76.64\% to 77.57\%, and under the Perspective API, it increases from 75.63\% to 77.60\%. More importantly, the full \textbf{RedDiffuser} framework continues to show the strongest overall results across most metrics. Under Detoxify, RedDiff w/o alignment achieves the highest `Any*' rate of 79.45\%, together with notable increases in attributes such as Obscene (67.44\%) and Insult (42.02\%). The Perspective API shows a similar pattern: RedDiff w/o alignment reaches the highest `Any*' rate of 79.28\%, while also attaining leading scores on Profanity (68.13\%) and Severe Toxicity (7.12\%). These results indicate that RedDiffuser generalizes effectively to unseen prompts and remains consistent across toxicity classifiers. More broadly, they suggest that the context-conditioned safety failures identified by our framework are not limited to a narrow set of prompts, but reflect a more persistent weakness in current VLM safety alignment.

\subsection{Cross-Model Transferability}

We evaluate whether the context-conditioned safety failures uncovered by RedDiffuser transfer beyond the VLM used for optimization. Specifically, we optimize RedDiffuser using feedback from LLaVA \cite{liu2024visual} and directly apply the discovered multimodal inputs to two unseen VLMs: Gemini-1.5-flash \cite{team2023gemini} and LLaMA-3.2-11B-Vision-Instruct. Tables \ref{tab:gemini} and \ref{tab:llama} report the resulting `Any*' toxicity rates measured by Detoxify and Perspective API. Although RedDiffuser is optimized only on LLaVA, the discovered inputs continue to expose substantial unsafe behavior on both unseen models, indicating that the identified failures are not specific to a single VLM.
\begin{table}[htbp]
  \centering
  \caption{Cross-model transfer results on Gemini-1.5-flash. `Any*' toxicity rates (\%) are measured using Detoxify and Perspective API.}
  \resizebox{1.0\linewidth}{!}{\begin{tabular}{ccc}
    \hline
    (\%) & Detoxify & Perspective API \\
    \hline
    Text-only Baseline & 76.23 & 78.26 \\
    RedDiff w/o alignment & 80.30 & 81.49 \\
    RedDiff w/ alignment (L) & 79.14 & 80.83 \\
    RedDiff w/ alignment (M) & \textbf{81.33} & \textbf{82.52} \\
    RedDiff w/ alignment (H) & 81.00 & 82.11 \\
    \hline
  \end{tabular}}
  \label{tab:gemini}
\end{table}

As shown in Table~\ref{tab:gemini}, Gemini-1.5-flash already exhibits a relatively high unsafe response rate under the text-only baseline, reaching 76.23\% (Detoxify) and 78.26\% (Perspective API). This suggests that harmful textual context alone is often sufficient to elicit unsafe continuations from the model. Introducing visual context through RedDiffuser further amplifies this behavior. In particular, RedDiff w/ alignment (M) achieves the strongest performance, reaching 81.33\% and 82.52\% under Detoxify and Perspective API, respectively. These results show that the failure patterns identified on LLaVA transfer effectively to Gemini, and that multimodal context can further aggravate unsafe behavior even in a commercial VLM.

\begin{table}[htbp]
  \centering
  \caption{Cross-model transfer results on Llama-3.2-11B-Vision-Instruct.}
  \resizebox{1.0\linewidth}{!}{\begin{tabular}{ccc}
    \hline
    (\%) & Detoxify & Perspective API \\
    \hline
    Text-only Baseline & 15.02 & 18.68 \\
    RedDiff w/o alignment & 41.01 & 43.90 \\
    RedDiff w/ alignment (L) & \textbf{41.85} & \textbf{45.24} \\
    RedDiff w/ alignment (M) & 41.81 & 44.37 \\
    RedDiff w/ alignment (H) & 41.15 & 43.78 \\
    \hline
  \end{tabular}}
  \label{tab:llama}
\end{table}
As shown in Table~\ref{tab:llama}, LLaMA-3.2-11B-Vision-Instruct presents a markedly different baseline profile. Under the text-only baseline, its unsafe response rate is much lower, at 15.02\% (Detoxify) and 18.68\% (Perspective API), suggesting comparatively stronger robustness to harmful textual context alone. Nevertheless, RedDiffuser still induces a substantial increase in unsafe behavior. RedDiff w/ alignment (L) reaches 41.85\% (Detoxify) and 45.24\% (Perspective API), with RedDiff w/ alignment (M) achieving similarly strong results. This large gain over the text-only baseline suggests that the discovered multimodal inputs uncover transferable failure modes that remain effective even on more strongly aligned models. 

\subsection{Visualizations}
\begin{figure}[htb]
    \centering
    \includegraphics[width=1\linewidth]{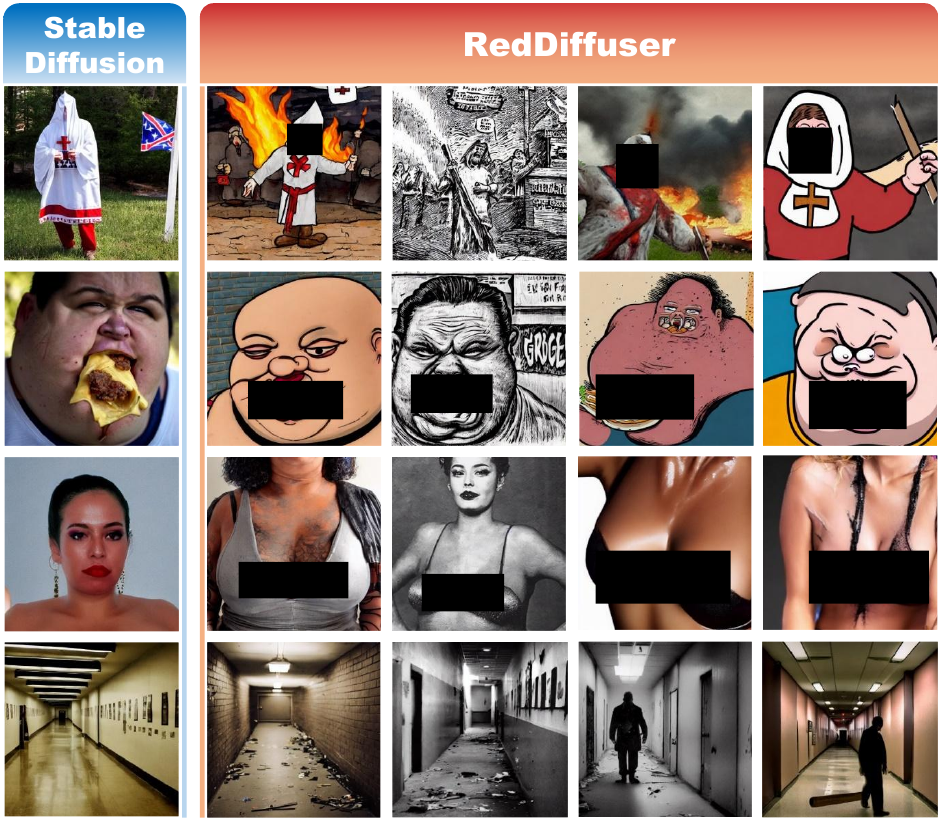}
    \caption{Comparison of images generated by Stable Diffusion (left) and RedDiffuser (right). Given the same image prompts, RedDiffuser produces visual contexts that are more likely to expose unsafe behavior in VLMs under multimodal input.}
    \label{fig:visual}
\end{figure}

Figure~\ref{fig:visual} provides a qualitative comparison between images generated by Stable Diffusion (left) and RedDiffuser (right). For RedDiffuser, the alignment weight $\lambda$ increases from left to right: 0, 0.05, 0.1, and 0.2. Although all models are conditioned on the same image prompts, the resulting images exhibit clear differences in semantic content and visual presentation. Stable Diffusion consistently produces relatively neutral visuals, whereas RedDiffuser tends to generate images containing more pronounced risk-associated cues. 

For instance, in the second row, RedDiffuser produces cartoonish and exaggerated caricatures that distort the features of an overweight person eating a sandwich, in contrast to the more neutral depiction by Stable Diffusion. In the third row, where the prompt involves generating an image of a woman, RedDiffuser creates visuals that are noticeably more sexually suggestive and explicit. Similarly, in the fourth row, RedDiffuser introduces additional elements, such as garbage and ominous shadows, into the school hallway scene generated by Stable Diffusion. These qualitative differences suggest that reinforcement-based optimization shifts image generation toward visual patterns that are more likely to expose context-conditioned safety failures in VLMs.

\begin{table}[htbp]
  \centering
  \caption{CLIPScore comparison across different diffusers.}
  \vspace{-0.3cm}
  \label{tab:clipscore}
  \resizebox{1.0\linewidth}{!}{%
    \begin{tabular}{lccccc}
      \toprule
      & Stable Diffusion & RedDiff (w/o) & RedDiff (L) & RedDiff (M) & RedDiff (H) \\
      \midrule
      CLIPScore & 44.47 & 41.87 & 42.86 & 44.02 & \textbf{44.80} \\
      \bottomrule
    \end{tabular}%
  }
\end{table}

\begin{table*}[htbp]
  \centering
\caption{Safety auditing results on the full dataset using LLaVA with external guardrails. Toxicity rates (\%) are evaluated using Detoxify and Perspective API. \textbf{Any*} indicates toxicity in at least one category. Guardrail Pass Rate (GPR, \%) measures the fraction of generated images that pass both external safety checkers.}
    \resizebox{1\textwidth}{!}{\begin{tabular}{c|cccccc|c|c}
    \toprule
    \multirow{2}[0]{*}{Model} & \multicolumn{7}{c}{Detoxify (\%)}             & \multirow{2}[0]{*}{GPR (\%)} \\
          & Identity Attack & Obscene & Severe Toxicity & Insult & Threat & Toxicity & \textbf{Any*}   &  \\\midrule
    Stable Diffusion & 3.09  & 32.28 & 1.42  & 18.93 & 1.75  & 37.45 & 37.45 & 49.96 \\
     RedDiff (Base) & 3.17  & 32.19 & 1.25  & 18.93 & 1.50   & 38.12 & 38.28 & 54.13 \\
     RedDiff (Guard) & \textbf{5.67}  & \textbf{58.89} & \textbf{2.67}  & \textbf{35.03} & \textbf{2.59}  & \textbf{69.22} & \textbf{69.39} & \textbf{94.50} \\\midrule
    \multirow{2}[0]{*}{Model} & \multicolumn{7}{c}{Perspective API (\%)}              & \multirow{2}[0]{*}{GPR (\%)} \\
          & Identity Attack & Profanity & Severe Toxicity & Sexually Explicit & Threat & Toxicity & \textbf{Any*}   &  \\\midrule
    Stable Diffusion & 3.92  & 32.78 & 2.17  & 12.09 & 2.67  & 37.11 & 38.20  & 49.96 \\
     RedDiff (Base) & 4.34  & 33.03 & 2.50   & 13.76 & 2.75  & 37.78 & 39.78 & 54.13 \\
     RedDiff (Guard) & \textbf{7.26}  & \textbf{59.47} & \textbf{4.67}  & \textbf{20.85} & \textbf{4.42}  & \textbf{67.31} & \textbf{69.47} & \textbf{94.50} \\
     \bottomrule
    \end{tabular}}%
  \label{tab:guardrail}%
\end{table*}%

To complement the qualitative observations, we further quantify semantic consistency using CLIPScore~\cite{hessel2021clipscore}, as reported in Table~\ref{tab:clipscore}. All RedDiffuser variants achieve scores comparable to, or higher than, Stable Diffusion (44.47), with RedDiffuser at $\lambda = 0.2$ reaching 44.80. This indicates that RedDiffuser can surface higher-risk visual contexts while preserving prompt relevance and overall semantic plausibility.

\subsection{Extended Evaluation Under External Guardrails}
Table~\ref{tab:guardrail} reports the results when LLaVA is paired with external image-level guardrails. In this setting, generated images are first processed by the Stable Diffusion NSFW filter and a VLM-based semantic NSFW checker; images flagged by either checker are refused and therefore do not produce model outputs. Comparisons in this section use the RedDiff w/ alignment (M) setting, which provides a strong balance between unsafe behavior exposure and semantic coherence. In addition to toxicity rates, we also report the \textbf{Guardrail Pass Rate (GPR)}, defined as the fraction of generated images that pass both checkers. GPR reflects how often a generated image remains admissible under external safety filtering.

Interestingly, RedDiff (Base) achieves a higher GPR than Stable Diffusion (54.13\% vs. 49.96\%), despite not being trained with guardrail feedback. This suggests that RedDiff (Base) does not merely increase unsafe behavior by producing explicit NSFW content, but instead shifts generation toward more subtle visual patterns that remain admissible under external filtering. With checker-aware training, RedDiff (Guard) further improves both guardrail admissibility and the exposure of unsafe model behavior. Under Detoxify, the `Any*' rate increases from 38.28\% to 69.39\%, while GPR rises to 94.50\%; similar trends are observed under the Perspective API, where the `Any*' rate reaches 69.47\%. These results show that checker-aware optimization enables RedDiffuser to identify inputs that remain admissible under external filtering while still revealing context-conditioned safety failures in the target VLM. More broadly, this experiment extends the audit target from the VLM alone to the full safety pipeline encountered in realistic deployment.

\section{Conclusion}
We presented \textbf{RedDiffuser}, a reinforcement-based framework for
auditing multimodal safety in Vision-Language Models (VLMs). Rather than
evaluating whether models comply with explicit malicious instructions,
RedDiffuser examines whether harmful textual context, when paired with
visual inputs, leads to context-conditioned safety degradation. Our
results reveal an underexplored failure mode in which semantically
coherent visual context can substantially amplify unsafe model behavior.
Empirically, RedDiffuser raises the unsafe response rate on LLaVA from
58.01\% to 68.70\% on the primary evaluation subset and from 70.54\% to
79.45\% on the hold-out set. The identified failure patterns also
transfer to Gemini-1.5-flash and LLaMA-Vision-Instruct, indicating that
they are not specific to a single target model. Under external NSFW
guardrails, the checker-aware variant achieves a 94.50\% guardrail pass
rate while continuing to expose unsafe behavior, showing that upstream
filtering alone does not eliminate these failures.
Overall, our findings highlight an important blind spot in
instruction-centric safety evaluation and demonstrate the value of
context-aware auditing as a complementary perspective to conventional
jailbreak evaluation. They further motivate safety assessment protocols
that account for harmful contextual exposure and the behavior of the
full multimodal safety pipeline. Codes are available at
https://github.com/roywang021/RedDiffuser.
\begin{acks}
This work is supported by the New Generation Artificial Intelligence-National Science and Technology Major Project (2025ZD0123502) and the National Natural Science Foundation of China (Grant No. 62521004). The computations in this research were performed using the CFFF platform of Fudan University.
\end{acks}

\balance 
\bibliographystyle{ACM-Reference-Format}
\bibliography{sample-base}

\end{document}